\documentclass[10pt,twocolumn,letterpaper]{article}

\usepackage[pagenumbers]{iccv}
\definecolor{iccvblue}{rgb}{0.21,0.49,0.74}
\usepackage[pagebackref,breaklinks,colorlinks,allcolors=iccvblue]{hyperref}
\usepackage{subcaption}
\usepackage{caption}
\usepackage[ruled,vlined]{algorithm2e}
\usepackage{graphicx}
\usepackage{animate}
\usepackage[symbol]{footmisc}

\def\authorBlock{
    Ruixiao Dong$^{1,2}$,
    Mengde Xu$^{2}$,
    Zigang Geng$^{1,2}$,
    Li Li$^{1}$,
    Han Hu$^{2}$,
    Shuyang Gu$^{2}\footnotemark[1]$\\
    $^{\text{1}}$University of Science and Technology of China \ \ \ $^{\text{2}}$Tencent Hunyuan Research \\
    {\tt\small \{dongruixiaoyx, zigang\}@mail.ustc.edu.cn, lil1@ustc.edu.cn}\\
    {\tt\small \{jaredsheaxu, cientgu\}@tencent.com}
    
}

\newtheorem{definition}{Definition}[section]

\newif\ifreview 
\newif\ifarxiv 
\newif\ifcamera 
\newif\ifrebuttal

\newcommand{\framework}{figs/framework_v3.pdf
}

\newcommand{\lossCurve}{
figs/loss_curve.pdf
}

\newcommand{\longImage}{%
    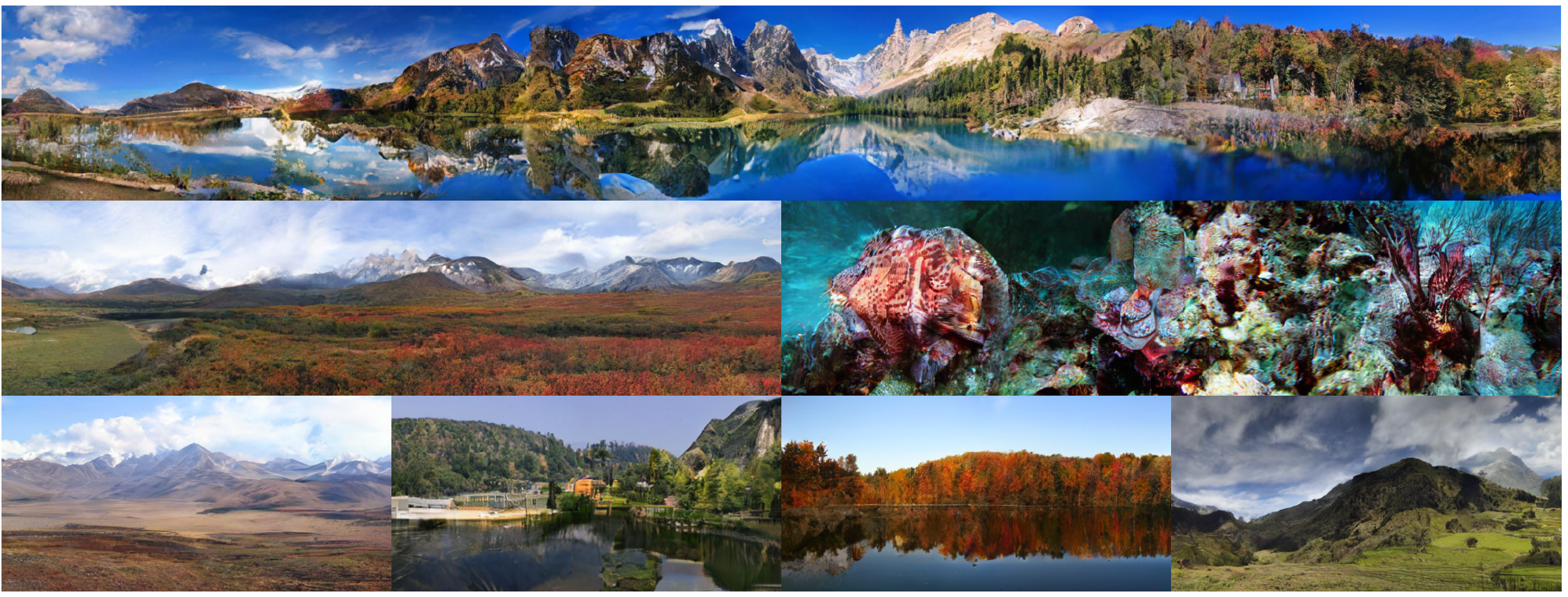
}

\title{Equivariant Image Modeling}
\author{\authorBlock}

\begin{document}
\maketitle
\renewcommand{\thefootnote}{\fnsymbol{footnote}}  
\footnotetext[1]{Corresponding Author.}

\begin{abstract}

Current generative models, such as autoregressive and diffusion approaches, decompose high-dimensional data distribution learning into a series of simpler subtasks. However, inherent conflicts arise during the joint optimization of these subtasks, and existing solutions fail to resolve such conflicts without sacrificing efficiency or scalability. We propose a novel equivariant image modeling framework that inherently aligns optimization targets across subtasks by leveraging the translation invariance of natural visual signals. Our method introduces (1) column-wise tokenization which enhances translational symmetry along the horizontal axis, and (2) windowed causal attention which enforces consistent contextual relationships across positions.
Evaluated on class-conditioned ImageNet generation at 256×256 resolution, our approach achieves performance comparable to state-of-the-art AR models while using fewer computational resources. Systematic analysis demonstrates that enhanced equivariance reduces inter-task conflicts, significantly improving zero-shot generalization and enabling ultra-long image synthesis. This work establishes the first framework for task-aligned decomposition in generative modeling, offering insights into efficient parameter sharing and conflict-free optimization. The code and models are publicly available at \url{https://github.com/drx-code/EquivariantModeling}.

\end{abstract}
\section{Introduction}
\label{sec:intro}

Generative modeling has gained significant attention in computer vision, particularly for image generation tasks. Recent advances in autoregressive (AR) models~\cite{esser2021taming, li2024autoregressive,tian2025visual} and diffusion processes~\cite{sohl2015deep, ho2020denoising, rombach2022high} demonstrate remarkable capabilities in modeling complex data distributions through a shared strategy: decomposing the challenging problem of learning high-dimensional distributions into sequences of simpler conditional distribution estimations. AR models achieve this through token-by-token prediction conditioned on previous outputs, while diffusion models employ iterative denoising across predefined noise levels. This multi-task learning paradigm raises critical questions about inter-task relationships - specifically, how decomposition strategies affect the synergies or conflicts between subtasks during joint optimization.

Recent investigations reveal inherent limitations in current decomposition frameworks. The MinSNR~\cite{hang2023efficient} identifies task conflicts in diffusion models and proposes Pareto optimization through careful loss weighting adjustments. An alternative approach, eDiff-I~\cite{balaji2022ediff} attempts to mitigate conflicts via task-specific parameter groups; however, this leads to an explosion in the number of parameters and ignores the relevance of different tasks. These methods fundamentally fail to address the conflicts in multi-task optimization. These observations motivate us to consider the core research question: \emph{Can we establish a principled task decomposition framework that inherently aligns optimization target across subtasks?}

In this paper, we present the first \emph{equivariant} image modeling paradigm that systematically minimizes inter-task conflicts. This task decomposition method is inspired by the unbounded visual signals in nature. When people look around, they do not perceive that the visual signals mutate at a fixed position. Therefore, we propose a left-to-right modeling framework that includes:

\emph{Equivariant} tokenization: We replace the conventional 2D patch grids with column-wise tokenization, enhancing spatial uniformity while better preserving natural image statistics (e.g., horizontal translation invariance in textures).

\emph{Equivariant} modeling: A windowed causal attention mechanism that enforces consistent contextual relationships across positions.

We validate our framework through comprehensive experiments on class-conditioned image generation. When evaluated on ImageNet at 256×256 resolution, our approach achieves performance comparable to state-of-the-art AR models while requiring substantially fewer computational resources. Through systematic analysis of the model's \emph{equivariance} properties, we demonstrate that enhanced task alignment significantly improves parameter sharing efficiency across subtasks - particularly benefiting zero-shot generalization capabilities. This intrinsic equivariance proves especially advantageous for generating unbounded natural scenes, outperforming human-collected datasets that contain spatial inductive bias.

Our principal contributions can be summarized as follows:
\begin{itemize}
\item The first \emph{equivariant} image modeling paradigm that fundamentally aligns subtask optimization target.
\item A column-wise 1D tokenization scheme that eliminates the spatial constraints inherent in conventional 2D grid-based approaches, delivering competitive performance on class-conditioned generation with fewer computational cost than standard AR models.
\item An analytical framework for quantifying subtask conflicts, with empirical evidence showing that enhanced equivariance improves zero-shot generalization and enables ultra-long image generation.
\end{itemize}
\section{Preliminaries}
\label{sec:prelim}
\subsection{Equivariance}
Modern image generation paradigms, including autoregressive~\cite{esser2021taming, tian2025visual} and diffusion models~\cite{ho2020denoising, rombach2022high}, decompose the complex task of image distribution modeling into explicit and traceable subtasks. This decomposition inherently formulates image generation as a multi-task learning problem. Considering that these different tasks share identical parameters, maintaining \emph{equivariance} among them becomes critical for learning efficiency~\cite{gu2024several}. Specifically, it is essential to ensure all subtasks exhibit congruent optimization trajectories under a shared parameter configuration. Formally, we define:

\begin{definition}
\label{def:equivariant }
 In a shared parameter space $\mathcal{H}$, given a set of subtasks $\{\mathcal{T}_{t}\}_{t=1}^{N}$ with their performance measurements $\{\mathcal{P}_{t}\}_{t=1}^{N}$, the subtask group is termed equivariant if for any pair $\mathcal{T}_i, \mathcal{T}_j \in \mathcal{T}$, their optimization directions coincide:
\begin{equation}
\theta^* = \underset{\theta \in \mathcal{H}}{\arg\max} \mathcal{P}_i(\theta) = \underset{\theta \in \mathcal{H}}{\arg\max} \mathcal{P}_j(\theta)
\end{equation}
where $\theta$ denotes the network parameters and $\mathcal{H}$ represents the optimization space.
\end{definition}

For instance, in image generation, if predicting pixels at different spatial locations constitutes distinct subtasks, equivariance implies that the optimization direction for predicting any pixel should remain consistent, regardless of its position. This property enables facilitates transfer learning across all spatial locations and efficient parameter sharing.

\subsection{Autoregressive Image Modeling}
\label{ssec:autoreg}
Autoregressive models decompose images into token sequences $\{x_1,...,x_t\}$ through the factorization:

\begin{equation}
\label{eq:ar}
p(x_{1},...,x_{n}|c)=\prod_{i=1}^{n}p(x_{i}|x_{<i},c),
\end{equation}
where $c$ represents the external condition, and each conditional distribution mapping $p(x_i|x_{< i},c)$ corresponds to a subtask predicting the $i$-th token given the preceding tokens $x_{<i}$ and the condition $c$. 

While enabling tractable likelihood estimation, conventional 2D grid-based autoregressive models face three critical challenges:

\begin{itemize}

\item \textbf{Subtask Heterogeneity}: The fixed raster-scan ordering~\cite{esser2021taming}\footnote{Though alternative orders like "Z order" exist, they exhibit similar subtask decomposition properties.} creates inherently distinct prediction tasks. For example, predicting border-region tokens typically proves more challenging than central-region tokens due to less local context.

\item \textbf{Non-stationary Context Dependence}: The standard autoregressive paradigm conditions each token generation on all preceding tokens, resulting in cumulatively increasing contextual dependencies. While wider context makes later positions easier to predict, it can cause the model to disproportionately optimize for these easier subtasks, potentially neglecting the more challenging predictions at earlier positions.

\item \textbf{Architectural Non-Equivariance}: Deep neural networks, particularly those employing self-attention mechanisms, can disrupt fundamental geometric symmetries. For example, position embeddings and attention patterns may introduce biases that violate translation invariance.

\end{itemize}

While recent work such as RAR~\cite{yu2024randomize} and TiTok~\cite{yu2024image} can partially address some challenges through stochastic ordering and 1D tokenization respectively, alleviating the inductive bias of AR image generation to a certain extent, there is still a lack of guarantee for the consistency of optimization directions across different tasks. This limitation potentially restricts their effectiveness in complex generation scenarios. Our work seeks to establish a subtask decomposition framework, where each subtask does not conflict and even promotes one another.

\section{Towards Equivariant Autoregressive Image Generation}
\label{sec:method}
\begin{figure*}
\centering
\includegraphics[width=\linewidth]{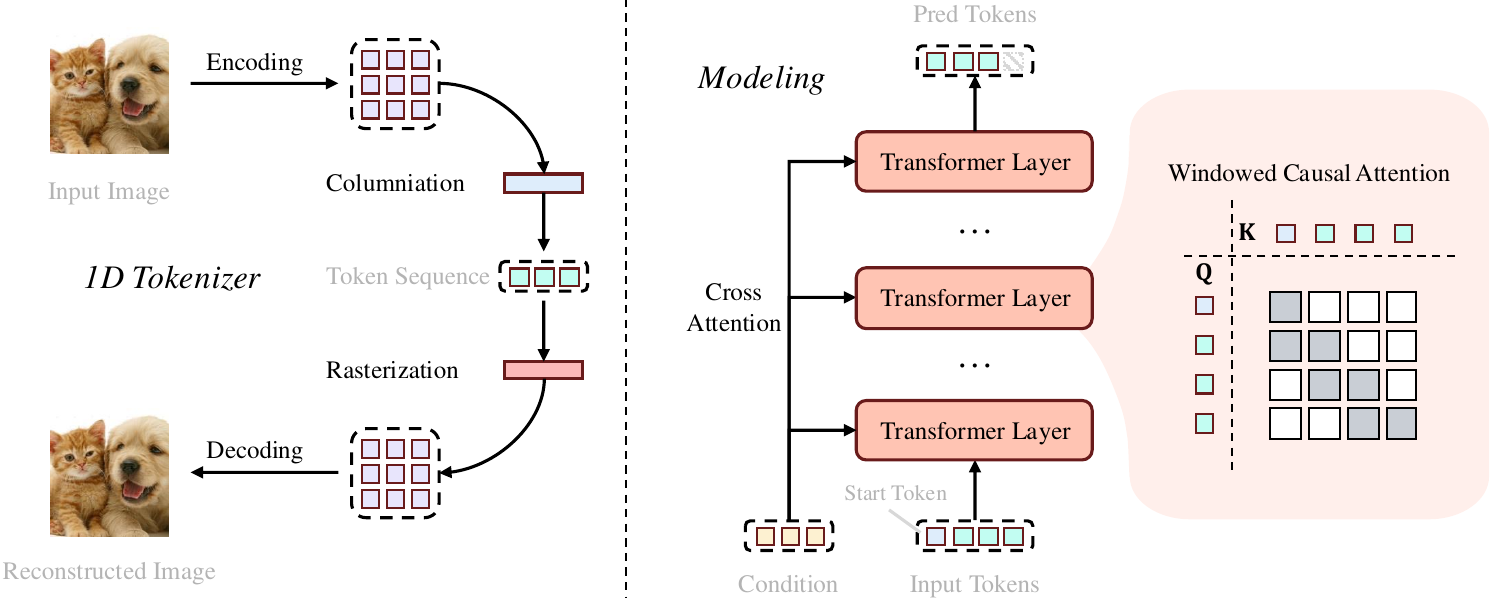}
\caption{\textbf{Illustration of Equivariant Image Generation Framework}. The tokenizer translates the image into 1D tokens arranged in columns and an enhanced autoregressive model models the column-wise token distribution.}
\label{fig:artch}
\end{figure*}

To address the challenges discussed in~\cref{ssec:autoreg}, we propose a column-based generation framework that explicitly removes the 2D grid structure and establishes a 1D equivariant modeling method, as illustrated in~\cref{fig:artch}. Our approach includes two key components: a column-wise tokenizer that naturally preserves vertical \emph{equivariance} by organizing image features into column-based 1D tokens (~\cref{sec:tokenizer}), and an autoregressive transformer equipped with \emph{equivariant} context (~\cref{sec:modeling}).

\subsection{Equivariant Tokenization via Columnization}
\label{sec:tokenizer}
\noindent\textbf{Columnization and Rasterization.} 
Visual signals in nature have no boundaries. When people look around, visual signals resemble a slowly unfolding scroll, without any sudden changes at fixed positions. Inspired by this, we consider tokenizing the image into a 1D latent sequence. Specifically, we employ \emph{columnization} to transform the classical 2D tokenizer into a 1D sequence. This process involves reshaping the height dimension into channels, followed by a linear projection to compress the representation into column-wise tokens. Specifically, given an output feature map of shape $H\times W\times C$ from a deep encoder, we transform it into a 1D token sequence $F$ of shape $W\times C'$ through:
\begin{equation}
\label{eq:col}
\begin{aligned}
H \times W\times C \xrightarrow[Reshape]{Permute} {W\times (HC)}\xrightarrow{Project} {W\times C'}.\
\end{aligned}
\end{equation}

\begin{figure}
    \centering
    \includegraphics[width=\linewidth]{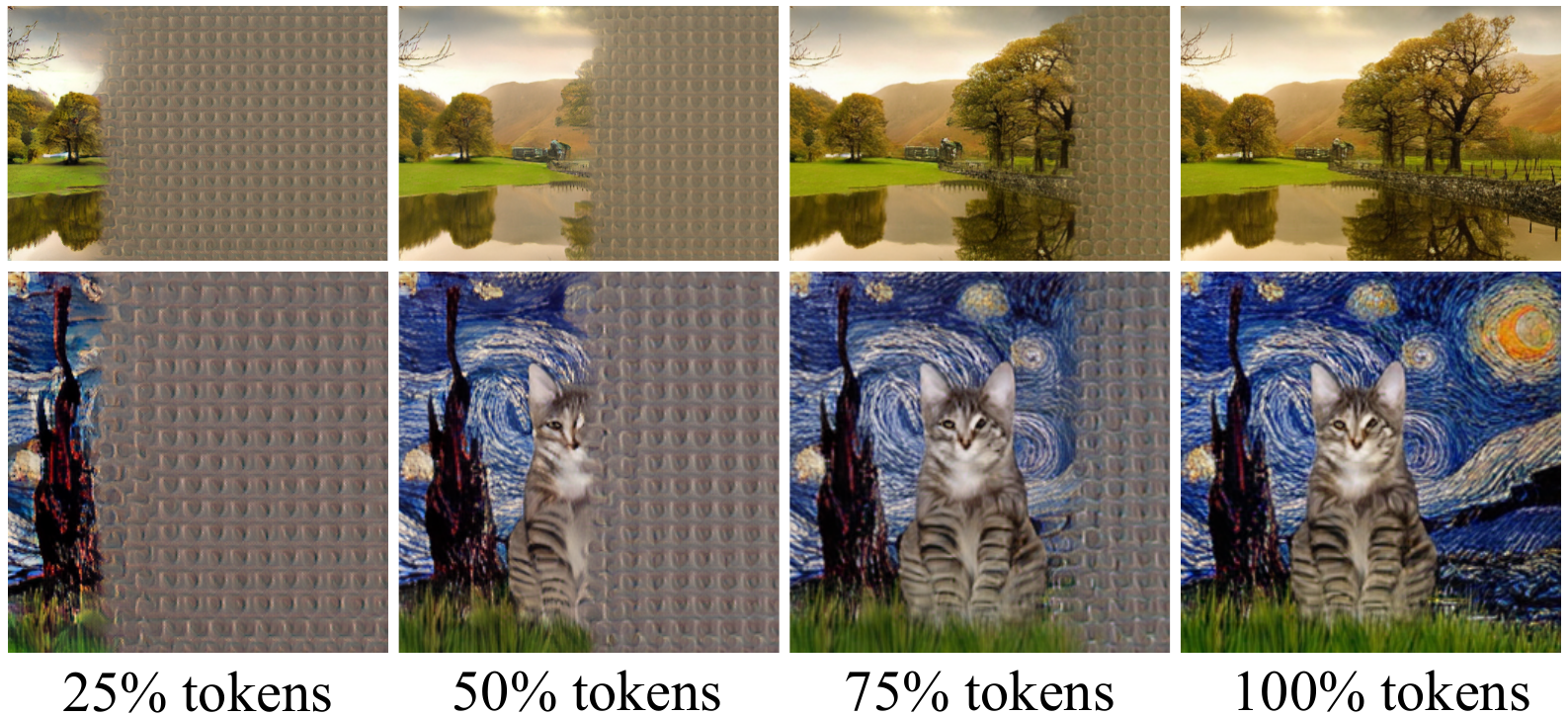}
    \caption{\textbf{Visual Meanings of 1D Tokens}. By progressively replacing the randomly initialized token sequence with tokens encoded from the ground truth images, the decoder faithfully reconstructs the original images step by step.}
    \label{fig:Spatial decoupling}
\end{figure}

This \emph{columnization} process eliminates the grid structure, resulting in each token representing a vertical visual signal, and adjacent tokens at any position transitioning naturally. Although the human-collected dataset may produce inductive biases due to camera settings and photographer preferences (e.g., the edge of the image may be darker due to the aperture), we leverage reflect padding to alleviate edge inconsistency. This semantically continuous transition between tokens provides the basic conditions for subsequent \emph{equivariant} autoregressive modeling. We visualize the relationship between tokens and visual content in~\cref{fig:Spatial decoupling}.

For image reconstruction, we rasterize the token sequence by projecting it to a higher-dimensional space and reshaping the channels back into the height dimension before processing it through a deep neural network:
\begin{equation}
\label{eq:raster}
\begin{aligned}
W\times C'\xrightarrow{Project}W\times (HC)\xrightarrow[Reshape]{Permute}
H\times W\times C. \
\end{aligned}
\end{equation}

\noindent\textbf{Semantic Aligned Tokenizer Training.}  Our \emph{columnization} and \emph{rasterization} operate on the deep features of an autoencoder inherited from~\cite{rombach2022high}. Following their approach, we train the tokenizer using multiple loss components: pixel-wise reconstruction loss $\mathcal{L}_{\text{rec}}$, adversarial loss $\mathcal{L}_{\text{gan}}$, perceptual loss $\mathcal{L}_{\text{p}}$, and KL divergence loss $\mathcal{L}_{\text{reg}}$ to regularize the token sequence distribution. Additionally, similar to~\cite{yu2024representation}, we introduce an alignment loss $\mathcal{L}_{\text{align}}$ that aligns the decoder's second-layer output features with those of a pretrained DINOv2~\cite{oquab2023dinov2} model, helping to preserve the latent space's semantic structure. The total loss combines the original autoencoder loss terms with semantic alignment loss:
\begin{equation}
\mathcal{L}_{\text{total}} = \lambda_1\mathcal{L}_{\text{rec}} + \lambda_2\mathcal{L}_{\text{reg}} + \lambda_3\mathcal{L}_{\text{p}} + \lambda_4\mathcal{L}_{\text{gan}} + \lambda_5\mathcal{L}_{\text{align}}
\end{equation}
where balancing coefficients ${\lambda_i}$ are set to 1.0, 0.01, 1.0, 0.5, and 5.0 by default.

\subsection{Equivariant Autoregressive Modeling}
\label{sec:modeling}

\noindent\textbf{Windowed Causal Attention.} As shown in~\cref{eq:ar}, autoregressive transformers employ causal attention to ensure that each token can attend to all its preceding tokens, making later subtasks significantly easier than earlier ones. A straightforward solution to address the potential imbalance in training is to limit each token's context to a fixed window of $k$ previous tokens:
\begin{equation}
\label{eq:window_ar}
p(x_{1},...,x_{n}|c)=\prod_{i=1}^{n}p(x_{i}|x_{\ge i-k,<i},c)
\end{equation}

We named this approach as \emph{Real Equivariant Modeling}. While this promotes \emph{equivariance} across subtasks, it compromises the long-range dependencies that are often useful for image generation~\cite{zhang2022styleswin}. As an alternative, we implement windowed causal attention with a fixed context size $w$ in each transformer layer. In this way, the network can increase the receptive field by stacking multiple layers, thereby implicitly modeling the long-range dependencies. For position $i$ in each layer, the attention operation becomes:
\begin{equation}
\text{Attn}(q_i, K, V) = \text{softmax}(\frac{q_iK_{i-w:i}^T}{\sqrt{d}})V_{i-w:j}
\end{equation}

where $q_i$ represents the query at position $i$, and $K_{i-w:i}, V_{i-w:i}$ denote key-value pairs from the previous $w$ positions. 
While this approach partially constrains the receptive field, empirical observations indicate that strategically leveraging the model's equivariant properties enhances generation quality. 

Our model architecture features a causal transformer with the windowed causal attention mechanism. In each self-attention layer, we utilize rotary position embedding~\cite{su2024roformer} to encode token positions within the sequence. To inject conditioning information, we place a cross-attention layer between the self-attention layer and feed-forward network in each transformer block. Previous works~\cite{zhang2022styleswin,hou2024high} find that the cross-attention requires absolute position embeddings to effectively convey global layout information. However, it may cause \emph{architectural non-equivariance}. Therefore, we mitigate this issue through position embedding augmentation, randomly shifting the position index of the entire sequence with a randomly selected value. Following previous work~\cite{li2024autoregressive}, we implement a diffusion head above the transformer to predict the next token from random noise. To accelerate training, we employ the flow matching algorithm~\cite{liu2022flow,lipman2022flow}, and the overall loss function is:
\begin{equation}
\mathcal{L} = E_{i\in[1,n], t}||\mathbf{D}(tx_{i}+(1-t)\epsilon_{i},t,z_i)-(x_i-\epsilon_{i})||^2,
\end{equation}
where $t$ denotes the noise level, which is sampled from $U(0,1)$. Accordingly, $x_i$ and $z_i$ denote the ground truth token and the output of the causal transformer at position $i$. $\epsilon_i$ is randomly sampled Gaussian noise, and $\mathbf{D}$ denotes the denoising network, which predicts the vector field at noise level $t$.

\section{Experiments}

\subsection{Experimental Settings}
We conduct experiments on the ImageNet-1k dataset~\cite{deng2009imagenet}. All images are resized to 256×256 resolution, utilizing standard augmentation techniques, including random cropping and horizontal flipping. We follow the common train/validation split to ensure consistent evaluation with prior approaches.

\noindent\textbf{Implementation Details.}
Our tokenizer architecture follows ~\cite{esser2021taming, rombach2022high}, employing a 2D encoder with a downsampling factor of 16, followed by a \emph{columniation} operation to produce token tensors $F'\in \mathbb{R}^{w\times c'}$ where $w=16, c'=256$. Training utilizes the Adam optimizer~\cite{kingma2014adam} for 320k iterations with a batch size of 192. The learning rate starts at $1.92\times 10^{-4}$, with a linear warm-up over 5,000 iterations and a 20\% decay every 30,000 iterations. The semantic loss is introduced after 20,000 iterations.

We train the generator for 1,200 epochs using AdamW~\cite{loshchilov2017decoupled} with a batch size of 2,048. The initial learning rate is $8\times 10^{-4}$, which is linearly warmed up over 100 epochs and then maintained at a constant level. The weight decay is set to 0.02, along with the momentum parameters $(\beta_1, \beta_2)=(0.9,0.95)$. An exponential moving average (EMA) of the parameters in the generator is maintained with a momentum of 0.9999.

\noindent\textbf{Evaluation Metrics.}
We evaluate reconstruction fidelity using the dataset-level reconstruction Fréchet Inception Distance (rFID)\cite{heusel2017gans}.  For generation quality, we measure both the generative Fréchet Inception Distance (gFID)\cite{heusel2017gans} and the Inception Score (IS)~\cite{salimans2016improved}.

\subsection{Deep Analysis on Equivariance}

Following the definition in~\cref{sec:prelim}, we examine the equivariance of our method by analyzing transferability across subtasks. We set up a MAR-AR variant model with 2D grid-based token sequences of length 16 as our baseline, denoted as AR-MAR-2D.

\begin{figure*}[htbp]
    \centering
    \includegraphics[width=\textwidth]{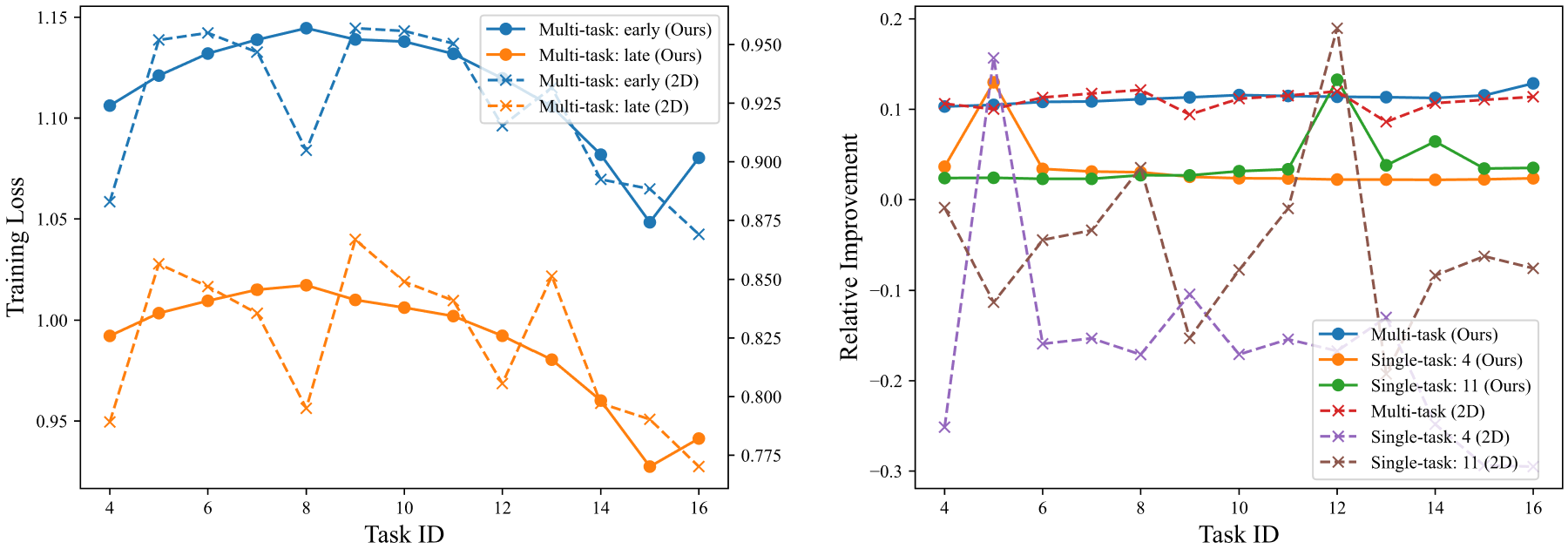}
    \caption{\textbf{Training Loss of Different Models}. Left: the training loss of different methods at early (10 epoches) and late (100 epoches) training stage. Right: the relative loss improvement of different methods under different settings compared to the early stage of \emph{Multi-task} setting. The higher value indicates better performance. The equivariant generation approach can transfer the improvement from a single task to other untrained tasks.}
    \label{fig:equivariant verification}
    \vspace{-0.3cm}
\end{figure*}

\noindent \textbf{Zero-shot Transfer between Subtasks.}
We investigate model generalization across different generative subtasks. With our tokenizer producing sequences of length 16, we have 16 distinct generative subtasks (detailed in \cref{ssec:autoreg}). We train the model on 8 selected subtasks and evaluate performance across all 16 subtasks using gFID. As shown in~\cref{tab:several_tasks}, while both approaches perform well when trained on all subtasks, our method maintains competitive performance under the zero-shot setting, with only a 3.42 point drop in gFID. In contrast, the 2D baseline struggles significantly with untrained subtasks, with gFID deteriorating from 7.93 to 92.46. This suggests our framework creates more consistent generative patterns across subtasks, enabling better generalization.

\begin{table}[tp]
\centering
\begin{tabular}{ccc}
\toprule
 Method & \# Num task. & gFID\\
\midrule
AR-MAR-2D & 16 & 7.93\\
Ours  & 16 & 5.57\\
\midrule
AR-MAR-2D & 8 & 92.46\\
Ours  & 8 & 8.99\\
\bottomrule
\end{tabular}
\caption{
\textbf{Performance under Zero-shot Setting.} \# Num task. is used to denote the number of trained subtasks. The total number of subtasks is 16 for all methods.
}
\vspace{-0.4cm}
\label{tab:several_tasks}
\end{table}

\noindent\textbf{Training Dynamics Analysis.}
We further examine \emph{equivariance} with training dynamics using two settings: 1) \textbf{Multi-task training} - the standard generative modeling setting, where a shared-parameter model is trained on all subtasks; 2) \textbf{Single-task training} - focusing on individual subtasks (4th and 11th) to isolate task interference effects.

~\cref{fig:equivariant verification} (left) shows the evolution of the multi-task training loss. While both methods converge to lower loss with longer training iterations, they exhibit different patterns across subtasks. The baseline consistently shows higher loss for subtasks corresponding to the image's boundary (5th, 9th, and 13th subtasks, which are the first tokens of each row in the 2D grid), while our method demonstrates more uniform behavior across subtasks, confirming better inter-task consistency. Notably, both methods show elevated loss for middle subtasks (corresponding to image centers), probably due to ImageNet's object-centric nature. We validates the data bias \begin{figure}[htbp]
    \centering
    \vspace{-1em}
    \includegraphics[width=\linewidth]{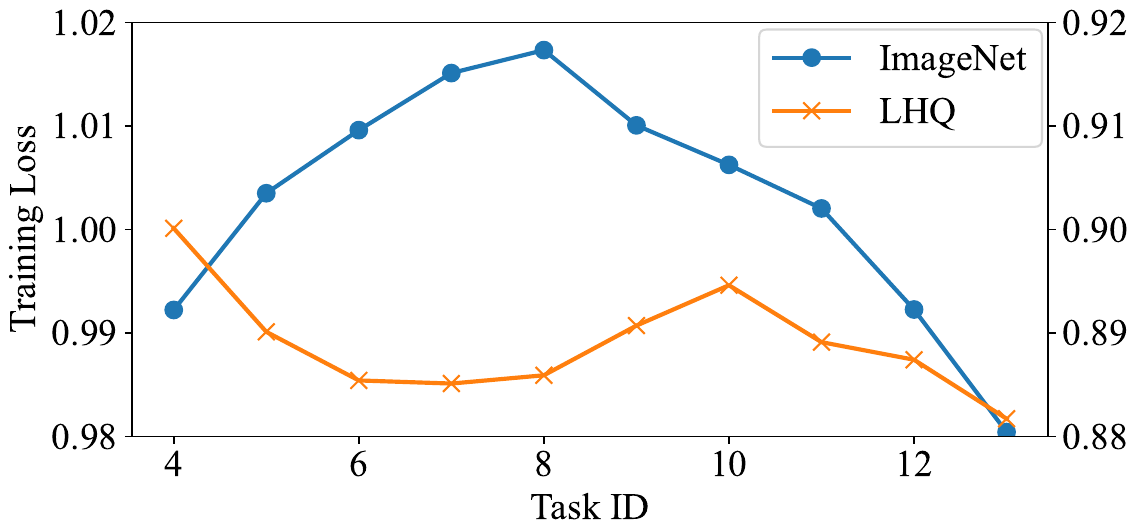}
    \caption{\textbf{Converged Training Loss on ImageNet vs LHQ}. Compared to ImageNet, the visual statics in LHQ demonstrates greater uniformity, as does the task-wise loss distribution.}
    \vspace{-0.3cm}
    \label{fig:data_diff}
\end{figure}
in~\cref{fig:data_diff} by comparing task-wise loss across different datasets. When testing on 
LHQ~\cite{skorokhodov2021aligning}, a landscape dataset with more uniform spatial distribution than ImageNet, the elevated middle loss phenomenon disappears. This motivates us to consider the dataset's influence in future research studying the equivariance problem.

Moreover, to isolate the influence of the data, we use the multi-task loss at the early training stage as our baseline and measure relative improvement across methods (with a lower loss indicating positive improvement). The results appear in the right panel of ~\cref{fig:equivariant verification}. In the multi-task setting, both our method and AR-MAR-2D converge to a 10\% relative improvement. However, in single-task training, our method successfully transfers the performance gain to all other subtasks, while the 2D baseline demonstrates negative impacts on untrained subtasks. These dynamics align with overall generation performance and suggest a promising approach for rapid \emph{equivariance} verification.

\begin{figure*}
    \centering
    \includegraphics[width=\linewidth]{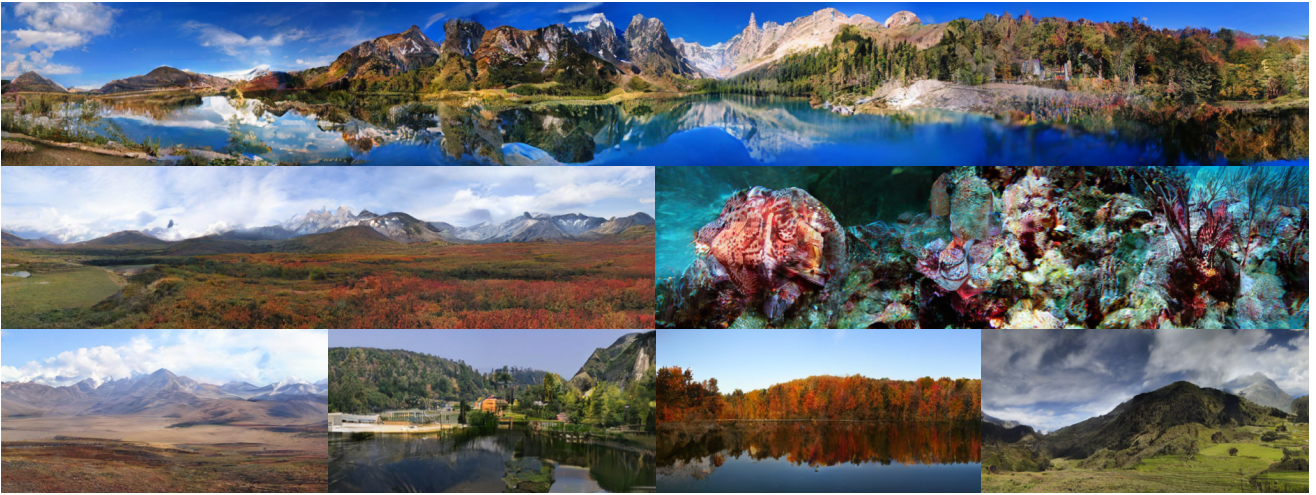}
    \vspace{-0.3cm}
    \caption{Visual examples of long image generation. We present visual examples of long images with arbitrary lengths, which are generated by our model that has been trained on the Places datasets with fixed length of 256.}
    \label{fig:Long Images}
    \vspace{-0.2cm}
\end{figure*}
\begin{figure}
    \centering
    \includegraphics[width=\linewidth]{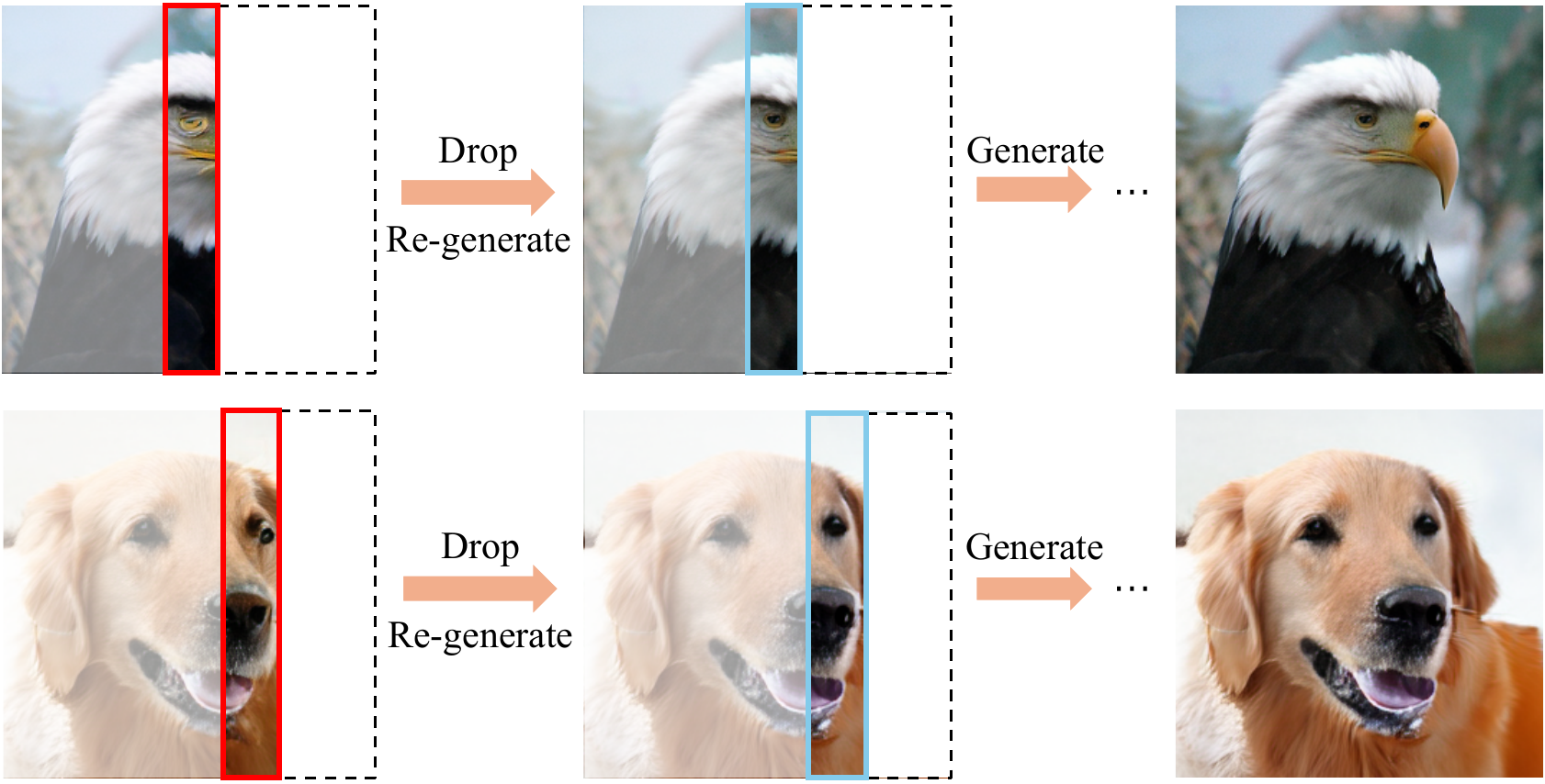}
    \caption{\textbf{Interactive Image Generation.} During inference, each token is immediately visible and bad generated tokens (circled with orange rectangle) are dropped according to human feedback.}
    \label{fig:instant_gen}
    \vspace{-0.2cm}
\end{figure}

\subsection{Equivariance Application}
\noindent\textbf{Long Content Image Generation.} Our experiments demonstrate that, due to the \emph{equivariance} of our method, the approach exhibits strong generalization ability across different subtasks. Inspired by the unbounded visual signals in nature, combining the generalization ability to generate training-unseen subtasks granted by the equivariant property, we test our models under long image generation scenarios. Specifically, we train our model on the Nature subset of the Places dataset~\cite{zhou2017places} for class-conditional generation, which contains 30 categories. The training image size is $256\times 256$, and we keep all other training parameters the same as in the ImageNet training phase. 

~\cref{fig:Long Images} showcases some generated examples of extended-length arbitrary resolution images produced by our model. These generated images exhibit a high spatial resolution, with lengths significantly greater than 256. The presented results demonstrably illustrate the zero-shot long-content capability of our method, primarily attributed to its inherent \emph{equivariance} property: although we only leverage $256\times 256$ resolution images to optimize the model, our approach effectively generates content at positional indices not encountered during training. Specifically, our method achieves the generation of images up to eight times longer than the input instances used during training, maintaining high visual fidelity and crucially avoiding discernible sharp edges between adjacent generated regions.

\noindent\textbf{Interactive Generation From Human Feedback.}
Another advantageous aspect of our method is that each token corresponds to a visually trackable area. This allows for interactive, token-wise feedback from users. We illustrate some examples in ~\cref{fig:instant_gen}. When a visually poor token is produced, we can promptly discard it and generate a new one. Ideally, our method could be integrated into an image editing pipeline, enabling complete control over the generation process. We leave this for future work.

\label{sec:experiments}

\begin{table}[tp]
\setlength{\tabcolsep}{4pt}
\renewcommand{\arraystretch}{1.05}
\begin{tabular}{ccccccc}
\toprule
Method &gFID$\downarrow$& IS$\uparrow$& \#Para. & \#Len. & GFLOPs$\downarrow$ \\
\midrule
\multicolumn{6}{c}{\textcolor{gray}{\emph{Diffusion}}}\\
DiT~\cite{peebles2023scalable} &  2.27 & 278.2 & 675M & -&118.64\\
\midrule
\multicolumn{6}{c}{\textcolor{gray}{\emph{MaskGIT}}}\\
TiTok~\cite{yu2024image}&  1.97 & 281.8 & 287M & 128 & 37.35 \\
MAR~\cite{li2024autoregressive} & 1.78 & 296.0 & 479M & 64 & 70.13\\
FractalMAR~\cite{li2025fractal} & 7.30 & 334.9 & 438M & - & 238.58\\

\midrule
\multicolumn{6}{c}{\textcolor{gray}{\emph{Autoregressive}}}\\
VQGAN~\cite{esser2021taming}&  15.78 & 74.3 &1.4B & 256 & 246.67 \\
VAR~\cite{tian2025visual}& 3.30 & 274.4 & 310M & 680 & 105.70\\
MAR~\cite{li2024autoregressive} &  4.69 & 244.6 & 479M & 64 & 78.50\\
\midrule
Ours-S &  7.21 & 233.70& 151M & 16 &  5.41\\
Ours-B & 5.57 & 260.05& 294M & 16 & 9.78\\
Ours-L & 4.48 & 259.91& 644M & 16 & 19.66\\
Ours-H & 4.17 & 290.66& 1.2B & 16 & 34.91\\
\bottomrule
\end{tabular}
\vspace{-0.1cm}
\caption{
\textbf{Class-conditional Generation Results on ImageNet 256×256 Benchmark.} \#Para. denotes the number of parameters in each generator, while \#Len. indicates the token sequence length that generators are required to model. 
}
\label{tab:modeling}
\end{table}
\subsection{System Comparison with SOTA Methods}
As detailed in ~\cref{tab:modeling}, a comparative analysis was conducted between our method and state-of-the-art generative methodologies, including diffusion models and autoregressive methods along with their variants. Our approach demonstrates comparable or better performance than the other methods presented in the table. Significantly, our model outperforms the autoregressive variant of MAR methods, confirming that the introduction of equivariant properties indeed enhances the modeling capability. Furthermore, the reduced token length inherent in our method leads to substantial computational savings in both training and inference, as evidenced by the reduction of GFLOPs in the table, particularly when compared to methods employing significantly longer token lengths ($\geq$ 64). 

We also compare our method with standard 2D variants in ~\cref{tab:fair_comparision}. With similar GFLOPs, our method attains superior generation performance. Moreover, our model achieves comparable gFID to the 256-token baseline while requiring only 15\% of the GFLOPs. The results demonstrate the effectiveness of our equivariance-driven design in achieving strong outcomes with minimal computational overhead.

\begin{table}[tp]
\centering
\renewcommand{\arraystretch}{1.05}
\begin{tabular}{ccccc}
\toprule
Method & Tokens Length&gFID$\downarrow$ & GFLOPs$\downarrow$\\
\midrule
AR-MAR-2D-B&256 & 3.99 &130.46\\
\midrule
AR-MAR-2D-B&16 & 7.93 &9.79\\
AR-MAR-2D-L & 16 & 7.49 & 19.68\\
Ours-B&16& 5.57 &9.78\\
Ours-L&16&4.48&19.66\\
\bottomrule
\end{tabular}
\caption{\textbf{Comparison on the performance and computational efficiency.} Our methods achieves better trade-off between performance and computation cost.}
\label{tab:fair_comparision}
\vspace{-0.2cm}
\end{table}

\subsection{Ablation Studies}

\noindent\textbf{Windowed Causal Attention.}  
To address the inconsistency in causal transformer models arising from subtasks adapting to varying context lengths, we introduce window-based attention. Specifically, each token is restricted to attending to a fixed-size local window of $\omega$ tokens during attention computation. To evaluate the impact of this technique, we perform comparative experiments on our small model with a window size of $\omega=3$ and compare it against standard causal attention, where each token attends to all preceding tokens. Additionally, we train a \textit{real equivariant} variant by restricting the context window length during training; in this scenario, the model only observes randomly cropped sequences of length $\omega$. Under these training conditions, the subsequences gain an explicit equivariant property: apart from the initial $\omega -1$ tokens, generated subsequences are invariant to shifted positions.

As summarized in ~\cref{tab:restricted_ctx}, limiting the receptive field slightly weakens the modeling of long-range dependencies but enhances consistency and uniformity across tasks, thereby improving the overall training of the generator. In particular, our generator with window causal attention surpasses the baseline model. Moreover, due to the fixed context size in the transformer layer, it reduces the theoretical computational cost of attention by approximately 42.9\% compared to causal attention. Conversely, while the \textit{real equivariant} generator maintains highly equivariant properties, it exhibits significantly reduced performance due to overly limited receptive fields, demonstrating that preserving long-range context information remains crucial for generative modeling.

\begin{figure}[t]
    \centering
    \begin{subfigure}[b]{0.47\textwidth}
        \includegraphics[width=\textwidth]{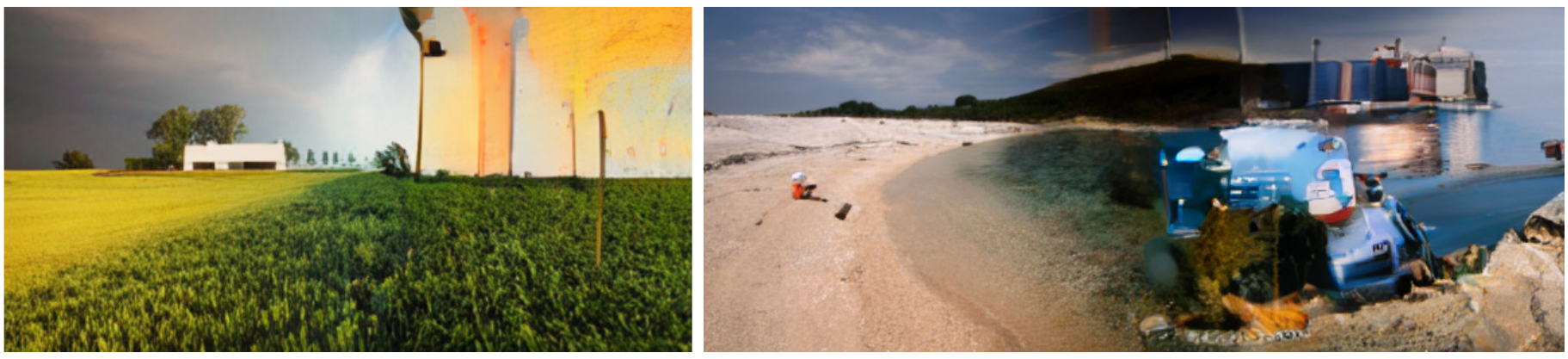}
         \vspace{-0.5cm}
        \caption{w/o random shift}
        \vspace{0.2cm}
    \end{subfigure}  
    \begin{subfigure}[b]{0.47\textwidth} 
        \includegraphics[width=\textwidth]{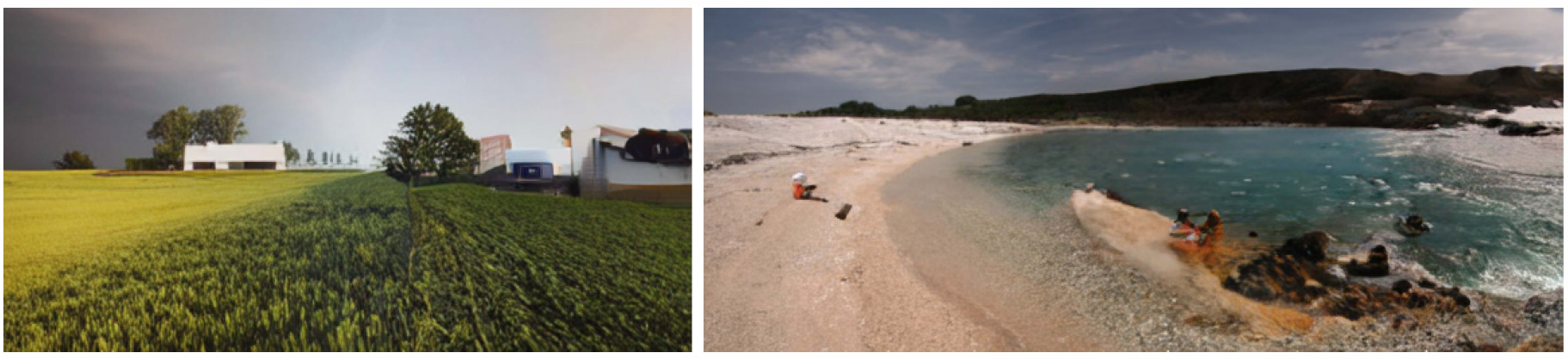}
        \vspace{-0.5cm}
        \caption{with random shift}
    \end{subfigure}
    \vspace{-0.1cm}
    \caption{\textbf{Visualization effect of the Augmented Position Embedding.} The model is trained on a subset of the Places dataset.}
    \label{fig:random_shift}
\end{figure}
\begin{table}[]
    \centering
    \renewcommand{\arraystretch}{1.05}
    \begin{tabular}{ccccc}
    \toprule
         Method&Equ.&L-Ctx.&gFID$\downarrow$ & Attn FLOPs$\downarrow$ \\
    \midrule
         Full Causal& & \checkmark & 7.35 & 4.2M\\
         Real Equ. & \checkmark & & 8.87& \textbf{0.26M}\\
         Window Causal &\checkmark&\checkmark& \textbf{7.21}& 1.8M\\
    \bottomrule
    \end{tabular}
    \vspace{-0.1cm}
    \caption{\textbf{Ablation Study of Windowed Causal Attention.} "Real Equ." indicates the strictly equivariant model variant as described in ~\cref{sec:modeling}; "Equ." refers to whether the model possesses equivariant properties; and "L-Ctx." denotes whether long-range contexts are utilized.}
    \label{tab:restricted_ctx}
    \vspace{-0.3cm}
\end{table}

\noindent\textbf{Position Embedding Augmentation.} We evaluate our proposed position embedding augmentation by training our large model on the Places dataset and comparing visualization results with and without this augmentation, which keeps the left side of the images as the original images, as illustrated in ~\cref{fig:random_shift}. We observe that the model trained without augmented position embeddings can still extrapolate beyond the training lengths to generate longer images; however, noticeable artifacts and discontinuities begin to appear in regions beyond the model's training positions. In contrast, when augmented position embeddings are utilized, the generated images exhibit stronger consistency across spatial locations, indicating that the model effectively generalizes beyond the observed spatial contexts. Specifically, as shown in the visual comparisons (~\cref{fig:random_shift}), our augmented model synthesizes coherent, artifact-free, and smooth images even at significantly increased lengths, demonstrating improved capability for extrapolation and robust generalization. 
\begin{table}[tp]
\centering
\begin{tabular}{lccc}
\toprule
&Method & rFID$\downarrow$ & gFID $\downarrow$\\
\midrule
1&baseline & 1.11 & 7.10\\
2&+Stronger discriminator & 0.62 & 6.29\\
3&+ Decoder finetune & 0.58 & 6.25\\
\textbf{Ours}&+Semantic aligned loss & 0.56 & 5.57\\
\bottomrule
\end{tabular}
\caption{
\textbf{Ablation on the Impact of Tokenizer Components.} "\textbf{Ours}" denotes the final setting we adopted in all other experiments.
}
\vspace{-0.1in}
\label{tab:Ablation1}
\end{table}

\noindent\textbf{Components in Tokenizer.} We analyze incremental improvements to our tokenizer within our base model, identified as (1) Baseline Tokenizer. The base tokenizer following LDM-tokenizer \cite{rombach2022high} achieves a gFID of 7.10 and an rFID of 1.11. (2) Enhanced Discriminator. Replacing the default discriminator with a DINO-small backbone \cite{zhang2022dino} improves rFID by 0.49 and gFID by 0.81, emphasizing stronger feature discrimination. (3) Decoder Fine-tuning. Fine-tuning the decoder enhances reconstruction fidelity and marginally improves generation quality. (4) Alignment Loss Introduction. Recognizing that the semantic information contained in the latent space is insufficient for high-quality generation, we introduce an alignment loss  $\mathcal{L}_{align}$ to align the latent representations with a pretrained DINOv2 model \cite{oquab2023dinov2}, resulting in a significant improvement in the gFID metric to 5.57.

\section{Related Work}
\label{sec:relateds}

Image distribution modeling is a long-standing topic in computer vision. One practical approach to solving this is visual signal decomposition, which breaks down the modeling task into subtasks.

Autoregressive models decompose each subtask as next-token generation. Early works like PixelCNN~\cite{van2016conditional,salimans2017pixelcnn++} decomposed image generation into pixel-by-pixel prediction in a raster-scan order. Later approaches~\cite{van2017neural, esser2021taming} introduced a two-stage process: first, encoding images into a latent space utilizing a tokenizer, and then employing autoregressive modeling to decompose latent modeling into sequential subtasks. While subsequent research~\cite{razavi2019generating, yu2021vector, lee2022autoregressive, zheng2022movq, mentzer2023finite, gu2024rethinking, takida2023hq, zhu2024scaling, zhu2024addressing, luo2024open, weber2024maskbit, chen2024deep, zhao2024epsilon, qu2024tokenflow} has focused on enhancing tokenizer capacity, the fundamental decomposition logic remains unchanged and continues to face challenges associated with 2D grid structures. Various works have been proposed to eliminate prediction on a 2D grid. VAR~\cite{tian2025visual} introduced scale-wise decomposition, predicting from small to large scales. However, this approach still exhibits significant non-equivariance, with earlier tokens primarily encoding low-frequency information and later tokens capturing high-frequency details. Concurrent work~\cite{ren2025next} proposes predicting block-wise or row-wise tokens in parallel, which is technically similar to our approach.

Diffusion models~\cite{sohl2015deep, ho2020denoising} take a different approach by decomposing visual signals into progressively denoised image sequences through shared-parameter models. Standard noise schedules, however, can create distributional inconsistencies across denoising stages. Several techniques have been proposed to address this challenge: the reparameterization trick~\cite{nichol2021improved, salimans2022progressive} has proven vital in unifying output distributions across various denoising tasks; Flow Matching~\cite{liu2022flow, lipman2022flow} and Consistency Models~\cite{song2023consistency} aim to develop improved noise strategies that maintain consistency throughout the denoising process; Min-SNR~\cite{tang2024simplified} employs Pareto optimization by carefully adjusting loss weights to identify conflicts among tasks; and Variational Diffusion Models~\cite{kingma2021variational} and the Diffusion Schrödinger Bridge~\cite{de2021diffusion, tang2024simplified} explore learning the noise-adding strategy rather than relying on fixed heuristics. Despite these advances, real-world application of these methods remains constrained by unresolved distributional mismatches.

MaskGIT~\cite{chang2022maskgit} decomposes the visual signals into depth-wise generations that sequentially predict sets of tokens. TikTok~\cite{yu2024image} further simplifies the decomposition using a 1D tokenizer. Muse~\cite{chang2023muse} alleviates subtask conflicts in depth-wise generation by dividing the subtasks into two groups and modeling them with distinct parameters.

Although numerous techniques have been proposed to properly decompose visual signals, our work is the first to address this problem from an \emph{equivariance} perspective, providing a more systematic analytical framework.

\label{sec:related}

\section{Conclusion}
\label{sec:conclusion}

This work establishes Equivariant Image Modeling as a principled framework for mitigating subtask conflicts in generative models. By introducing column-wise tokenization and equivariant token modeling, we demonstrate that the spatial equivariant decomposition aligns optimization targets, enhancing parameter efficiency and zero-shot generalization. Experimental results on ImageNet-1k generation validate the computational advantages of our approach over conventional autoregressive models. The exploration of equivariant task decomposition opens new directions for the future development of generative models.

{\small
\bibliographystyle{ieee_fullname}
\bibliography{main}

\begin{thebibliography}{10}\itemsep=-1pt

\bibitem{balaji2022ediff}
Yogesh Balaji, Seungjun Nah, Xun Huang, Arash Vahdat, Jiaming Song, Qinsheng Zhang, Karsten Kreis, Miika Aittala, Timo Aila, Samuli Laine, et~al.
\newblock ediff-i: Text-to-image diffusion models with an ensemble of expert denoisers.
\newblock {\em arXiv preprint arXiv:2211.01324}, 2022.

\bibitem{chang2023muse}
Huiwen Chang, Han Zhang, Jarred Barber, Aaron Maschinot, Jose Lezama, Lu Jiang, Ming-Hsuan Yang, Kevin~Patrick Murphy, William~T Freeman, Michael Rubinstein, et~al.
\newblock Muse: Text-to-image generation via masked generative transformers.
\newblock In {\em ICML}, pages 4055--4075. PMLR, 2023.

\bibitem{chang2022maskgit}
Huiwen Chang, Han Zhang, Lu Jiang, Ce Liu, and William~T Freeman.
\newblock Maskgit: Masked generative image transformer.
\newblock In {\em CVPR}, pages 11315--11325, 2022.

\bibitem{chen2024deep}
Junyu Chen, Han Cai, Junsong Chen, Enze Xie, Shang Yang, Haotian Tang, Muyang Li, and Song Han.
\newblock Deep compression autoencoder for efficient high-resolution diffusion models.
\newblock In {\em ICLR}, 2025.

\bibitem{de2021diffusion}
Valentin De~Bortoli, James Thornton, Jeremy Heng, and Arnaud Doucet.
\newblock Diffusion schr{\"o}dinger bridge with applications to score-based generative modeling.
\newblock {\em NeurIPS}, 34:17695--17709, 2021.

\bibitem{deng2009imagenet}
Jia Deng, Wei Dong, Richard Socher, Li-Jia Li, Kai Li, and Li Fei-Fei.
\newblock Imagenet: A large-scale hierarchical image database.
\newblock In {\em CVPR}, pages 248--255, 2009.

\bibitem{esser2021taming}
Patrick Esser, Robin Rombach, and Bjorn Ommer.
\newblock Taming transformers for high-resolution image synthesis.
\newblock In {\em CVPR}, 2021.

\bibitem{gu2024several}
Shuyang Gu.
\newblock Several questions of visual generation in 2024.
\newblock {\em arXiv preprint arXiv:2407.18290}, 2024.

\bibitem{gu2024rethinking}
Yuchao Gu, Xintao Wang, Yixiao Ge, Ying Shan, and Mike~Zheng Shou.
\newblock Rethinking the objectives of vector-quantized tokenizers for image synthesis.
\newblock In {\em CVPR}, pages 7631--7640, 2024.

\bibitem{hang2023efficient}
Tiankai Hang, Shuyang Gu, Chen Li, Jianmin Bao, Dong Chen, Han Hu, Xin Geng, and Baining Guo.
\newblock Efficient diffusion training via min-snr weighting strategy.
\newblock In {\em ICCV}, pages 7441--7451, 2023.

\bibitem{heusel2017gans}
Martin Heusel, Hubert Ramsauer, Thomas Unterthiner, Bernhard Nessler, and Sepp Hochreiter.
\newblock Gans trained by a two time-scale update rule converge to a local nash equilibrium.
\newblock {\em NeurIPS}, 30, 2017.

\bibitem{ho2020denoising}
Jonathan Ho, Ajay Jain, and Pieter Abbeel.
\newblock Denoising diffusion probabilistic models.
\newblock {\em NeurIPS}, 33:6840--6851, 2020.

\bibitem{hou2024high}
Runtong Hou and Xu Zhao.
\newblock High-quality talking face generation via cross-attention transformer.
\newblock In {\em 2024 IEEE International Conference on Real-time Computing and Robotics (RCAR)}, pages 194--199, 2024.

\bibitem{kingma2021variational}
Diederik Kingma, Tim Salimans, Ben Poole, and Jonathan Ho.
\newblock Variational diffusion models.
\newblock {\em NeurIPS}, 34:21696--21707, 2021.

\bibitem{kingma2014adam}
Diederik~P. Kingma and Jimmy Ba.
\newblock Adam: A method for stochastic optimization.
\newblock In {\em ICLR}, 2015.

\bibitem{lee2022autoregressive}
Doyup Lee, Chiheon Kim, Saehoon Kim, Minsu Cho, and Wook-Shin Han.
\newblock Autoregressive image generation using residual quantization.
\newblock In {\em CVPR}, pages 11523--11532, 2022.

\bibitem{li2025fractal}
Tianhong Li, Qinyi Sun, Lijie Fan, and Kaiming He.
\newblock Fractal generative models.
\newblock {\em arXiv preprint arXiv:2502.17437}, 2025.

\bibitem{li2024autoregressive}
Tianhong Li, Yonglong Tian, He Li, Mingyang Deng, and Kaiming He.
\newblock Autoregressive image generation without vector quantization.
\newblock {\em NeurIPS}, 37:56424--56445, 2024.

\bibitem{lipman2022flow}
Yaron Lipman, Ricky T.~Q. Chen, Heli Ben-Hamu, Maximilian Nickel, and Matthew Le.
\newblock Flow matching for generative modeling.
\newblock In {\em ICLR}, 2023.

\bibitem{liu2022flow}
Xingchao Liu, Chengyue Gong, and qiang liu.
\newblock Flow straight and fast: Learning to generate and transfer data with rectified flow.
\newblock In {\em ICLR}, 2023.

\bibitem{loshchilov2017decoupled}
Ilya Loshchilov and Frank Hutter.
\newblock Decoupled weight decay regularization.
\newblock In {\em ICLR}, 2019.

\bibitem{luo2024open}
Zhuoyan Luo, Fengyuan Shi, Yixiao Ge, Yujiu Yang, Limin Wang, and Ying Shan.
\newblock Open-magvit2: An open-source project toward democratizing auto-regressive visual generation.
\newblock {\em arXiv preprint arXiv:2409.04410}, 2024.

\bibitem{mentzer2023finite}
Fabian Mentzer, David Minnen, Eirikur Agustsson, and Michael Tschannen.
\newblock Finite scalar quantization: {VQ}-{VAE} made simple.
\newblock In {\em ICLR}, 2024.

\bibitem{nichol2021improved}
Alexander~Quinn Nichol and Prafulla Dhariwal.
\newblock Improved denoising diffusion probabilistic models.
\newblock In {\em ICML}, pages 8162--8171. PMLR, 2021.

\bibitem{oquab2023dinov2}
Maxime Oquab, Timoth{\'e}e Darcet, Th{\'e}o Moutakanni, Huy~V. Vo, Marc Szafraniec, Vasil Khalidov, Pierre Fernandez, Daniel HAZIZA, Francisco Massa, Alaaeldin El-Nouby, Mido Assran, Nicolas Ballas, Wojciech Galuba, Russell Howes, Po-Yao Huang, Shang-Wen Li, Ishan Misra, Michael Rabbat, Vasu Sharma, Gabriel Synnaeve, Hu Xu, Herve Jegou, Julien Mairal, Patrick Labatut, Armand Joulin, and Piotr Bojanowski.
\newblock {DINO}v2: Learning robust visual features without supervision.
\newblock {\em Transactions on Machine Learning Research}, 2024.

\bibitem{peebles2023scalable}
William Peebles and Saining Xie.
\newblock Scalable diffusion models with transformers.
\newblock In {\em ICCV}, pages 4195--4205, 2023.

\bibitem{qu2024tokenflow}
Liao Qu, Huichao Zhang, Yiheng Liu, Xu Wang, Yi Jiang, Yiming Gao, Hu Ye, Daniel~K Du, Zehuan Yuan, and Xinglong Wu.
\newblock Tokenflow: Unified image tokenizer for multimodal understanding and generation.
\newblock {\em arXiv preprint arXiv:2412.03069}, 2024.

\bibitem{razavi2019generating}
Ali Razavi, Aaron Van~den Oord, and Oriol Vinyals.
\newblock Generating diverse high-fidelity images with vq-vae-2.
\newblock {\em NeurIPS}, 32, 2019.

\bibitem{ren2025next}
Shuhuai Ren, Shuming Ma, Xu Sun, and Furu Wei.
\newblock Next block prediction: Video generation via semi-auto-regressive modeling.
\newblock {\em arXiv preprint arXiv:2502.07737}, 2025.

\bibitem{rombach2022high}
Robin Rombach, Andreas Blattmann, Dominik Lorenz, Patrick Esser, and Bj{\"o}rn Ommer.
\newblock High-resolution image synthesis with latent diffusion models.
\newblock In {\em CVPR}, 2022.

\bibitem{salimans2016improved}
Tim Salimans, Ian Goodfellow, Wojciech Zaremba, Vicki Cheung, Alec Radford, and Xi Chen.
\newblock Improved techniques for training gans.
\newblock {\em NeurIPS}, 29, 2016.

\bibitem{salimans2022progressive}
Tim Salimans and Jonathan Ho.
\newblock Progressive distillation for fast sampling of diffusion models.
\newblock In {\em ICLR}, 2022.

\bibitem{salimans2017pixelcnn++}
Tim Salimans, Andrej Karpathy, Xi Chen, and Diederik~P Kingma.
\newblock Pixelcnn++: Improving the pixelcnn with discretized logistic mixture likelihood and other modifications.
\newblock {\em arXiv preprint arXiv:1701.05517}, 2017.

\bibitem{skorokhodov2021aligning}
Ivan Skorokhodov, Grigorii Sotnikov, and Mohamed Elhoseiny.
\newblock Aligning latent and image spaces to connect the unconnectable.
\newblock In {\em ICCV}, pages 14144--14153, 2021.

\bibitem{sohl2015deep}
Jascha Sohl-Dickstein, Eric Weiss, Niru Maheswaranathan, and Surya Ganguli.
\newblock Deep unsupervised learning using nonequilibrium thermodynamics.
\newblock In {\em ICML}, pages 2256--2265. PMLR, 2015.

\bibitem{song2023consistency}
Yang Song, Prafulla Dhariwal, Mark Chen, and Ilya Sutskever.
\newblock Consistency models.
\newblock In {\em ICML}, pages 32211--32252. PMLR, 2023.

\bibitem{su2024roformer}
Jianlin Su, Murtadha Ahmed, Yu Lu, Shengfeng Pan, Wen Bo, and Yunfeng Liu.
\newblock Roformer: Enhanced transformer with rotary position embedding.
\newblock {\em Neurocomputing}, 568:127063, 2024.

\bibitem{takida2023hq}
Yuhta Takida, Yukara Ikemiya, Takashi Shibuya, Kazuki Shimada, Woosung Choi, Chieh-Hsin Lai, Naoki Murata, Toshimitsu Uesaka, Kengo Uchida, Wei-Hsiang Liao, and Yuki Mitsufuji.
\newblock {HQ}-{VAE}: Hierarchical discrete representation learning with variational bayes.
\newblock {\em Transactions on Machine Learning Research}, 2024.

\bibitem{tang2024simplified}
Zhicong Tang, Tiankai Hang, Shuyang Gu, Dong Chen, and Baining Guo.
\newblock Simplified diffusion schr$\backslash$" odinger bridge.
\newblock {\em arXiv preprint arXiv:2403.14623}, 2024.

\bibitem{tian2025visual}
Keyu Tian, Yi Jiang, Zehuan Yuan, Bingyue Peng, and Liwei Wang.
\newblock Visual autoregressive modeling: Scalable image generation via next-scale prediction.
\newblock {\em NeurIPS}, 37:84839--84865, 2025.

\bibitem{van2016conditional}
Aaron Van~den Oord, Nal Kalchbrenner, Lasse Espeholt, Oriol Vinyals, Alex Graves, et~al.
\newblock Conditional image generation with pixelcnn decoders.
\newblock {\em NeurIPS}, 29, 2016.

\bibitem{van2017neural}
Aaron Van Den~Oord, Oriol Vinyals, et~al.
\newblock Neural discrete representation learning.
\newblock {\em NIPS}, 30, 2017.

\bibitem{weber2024maskbit}
Mark Weber, Lijun Yu, Qihang Yu, Xueqing Deng, Xiaohui Shen, Daniel Cremers, and Liang-Chieh Chen.
\newblock Maskbit: Embedding-free image generation via bit tokens.
\newblock {\em Transactions on Machine Learning Research}, 2024.

\bibitem{yu2021vector}
Jiahui Yu, Xin Li, Jing~Yu Koh, Han Zhang, Ruoming Pang, James Qin, Alexander Ku, Yuanzhong Xu, Jason Baldridge, and Yonghui Wu.
\newblock Vector-quantized image modeling with improved {VQGAN}.
\newblock In {\em ICLR}, 2022.

\bibitem{yu2024randomize}
Qihang Yu, Ju He, Xueqing Deng, Xiaohui Shen, and Liang-Chieh Chen.
\newblock Randomized autoregressive visual generation.
\newblock {\em arXiv preprint arxiv: 2411.00776}, 2024.

\bibitem{yu2024image}
Qihang Yu, Mark Weber, Xueqing Deng, Xiaohui Shen, Daniel Cremers, and Liang-Chieh Chen.
\newblock An image is worth 32 tokens for reconstruction and generation.
\newblock {\em NeurIPS}, 37:128940--128966, 2024.

\bibitem{yu2024representation}
Sihyun Yu, Sangkyung Kwak, Huiwon Jang, Jongheon Jeong, Jonathan Huang, Jinwoo Shin, and Saining Xie.
\newblock Representation alignment for generation: Training diffusion transformers is easier than you think.
\newblock In {\em ICLR}, 2025.

\bibitem{zhang2022styleswin}
Bowen Zhang, Shuyang Gu, Bo Zhang, Jianmin Bao, Dong Chen, Fang Wen, Yong Wang, and Baining Guo.
\newblock Styleswin: Transformer-based gan for high-resolution image generation.
\newblock In {\em CVPR}, pages 11304--11314, 2022.

\bibitem{zhang2022dino}
Hao Zhang, Feng Li, Shilong Liu, Lei Zhang, Hang Su, Jun Zhu, Lionel Ni, and Heung-Yeung Shum.
\newblock {DINO}: {DETR} with improved denoising anchor boxes for end-to-end object detection.
\newblock In {\em ICLR}, 2023.

\bibitem{zhao2024epsilon}
Long Zhao, Sanghyun Woo, Ziyu Wan, Yandong Li, Han Zhang, Boqing Gong, Hartwig Adam, Xuhui Jia, and Ting Liu.
\newblock $\epsilon$-vae: Denoising as visual decoding.
\newblock {\em arXiv preprint arXiv:2410.04081}, 2024.

\bibitem{zheng2022movq}
Chuanxia Zheng, Tung-Long Vuong, Jianfei Cai, and Dinh Phung.
\newblock Movq: Modulating quantized vectors for high-fidelity image generation.
\newblock {\em NeurIPS}, 35:23412--23425, 2022.

\bibitem{zhou2017places}
Bolei Zhou, Agata Lapedriza, Aditya Khosla, Aude Oliva, and Antonio Torralba.
\newblock Places: A 10 million image database for scene recognition.
\newblock {\em IEEE TPAMI}, 2017.

\bibitem{zhu2024scaling}
Lei Zhu, Fangyun Wei, Yanye Lu, and Dong Chen.
\newblock Scaling the codebook size of {VQ}-{GAN} to 100,000 with a utilization rate of 99\%.
\newblock In {\em NeurIPS}, 2024.

\bibitem{zhu2024addressing}
Yongxin Zhu, Bocheng Li, Yifei Xin, and Linli Xu.
\newblock Addressing representation collapse in vector quantized models with one linear layer.
\newblock {\em arXiv preprint arXiv:2411.02038}, 2024.

\end{thebibliography}
}

\newpage
\appendix
\label{sec:appendix}

\section{Datasets}\label{sec:datastes} 

\textbf{ImageNet-1k} \quad ImageNet~\cite{deng2009imagenet} is a large-scale hierarchical image database that has served as a cornerstone dataset for modern computer vision research since its introduction in 2009. It contains more than one million annotated images across 1,000 object categories, providing a robust benchmark for numerous vision tasks, including image classification, object detection, and class-conditional image generation. 

Dataset website: \url{https://image-net.org/}

\noindent\textbf{Places} \quad The Places~\cite{zhou2017places} dataset is curated according to the principles of human visual cognition, with the aim of creating a comprehensive resource to train artificial systems in high-level visual understanding tasks. Applications range from scene recognition and object detection in contextual environments, to sophisticated understanding tasks such as action recognition, event prediction, and theory-of-mind inference. The entire Places database includes more than 10 million images, covering over 400 unique scene categories. In particular, for our long-image generation experiments, we selected 30 nature categories from the Places-Challenge subset, which contains approximately 1 million images.

Dataset website: \url{http://places2.csail.mit.edu/}

\section{Implementation Details}\label{sec:implement_details}

\subsection{Data Augmentation}
We perform data augmentation by initially resizing the input images so that the smaller dimension is 256 pixels. Following this, random cropping is applied to the resized images. Additionally, horizontal flipping is performed with a probability of 0.5 to improve the robustness and generalization of the model.

\begin{table}[tp]
\centering
\begin{tabular}{lc}
\toprule
config &  value \\
\midrule
Base channels & 128 \\
Base channel multiplier per stage & [1, 1, 2, 4, 4] \\
Residual blocks per stage & 2 \\
Attention resolutions & 16 \\
Token channels & 256 \\
\midrule
Adversarial loss enabled at iteration & 5000 \\
Discriminator loss weight & 0.5 \\
Discriminator loss & hinge loss \\
Perceptual loss weight & 1.0 \\
Semantic anlignment loss enabled at iteration & 20000 \\
Semantic anlignment loss weight & 5.0 \\
KL divergence loss weight & 0.01 \\
\midrule
Gradient clipping by norm & 1.0 \\
Optimizer & Adam \\
Beta1 & 0.5 \\
Beta2 & 0.9 \\
Base LR & 1.92e-4 \\
LR warmup iterations & 5000 \\
LR decay frequency & 30000 \\
LR decay ratio & 0.2 \\
\midrule
EMA decay & 0.9999 \\ 
\midrule
Training epochs & 50 \\
Total Batchsize & 192 \\
GPU & A100 \\
\bottomrule
\end{tabular}
\caption{Detailed hyper-parameters for our equivariant 1D tokenizer.}
\label{tab:stage1}
\vspace{-0.6cm}
\end{table}

\begin{figure}[tp]
    \centering
    \includegraphics[width=\linewidth]{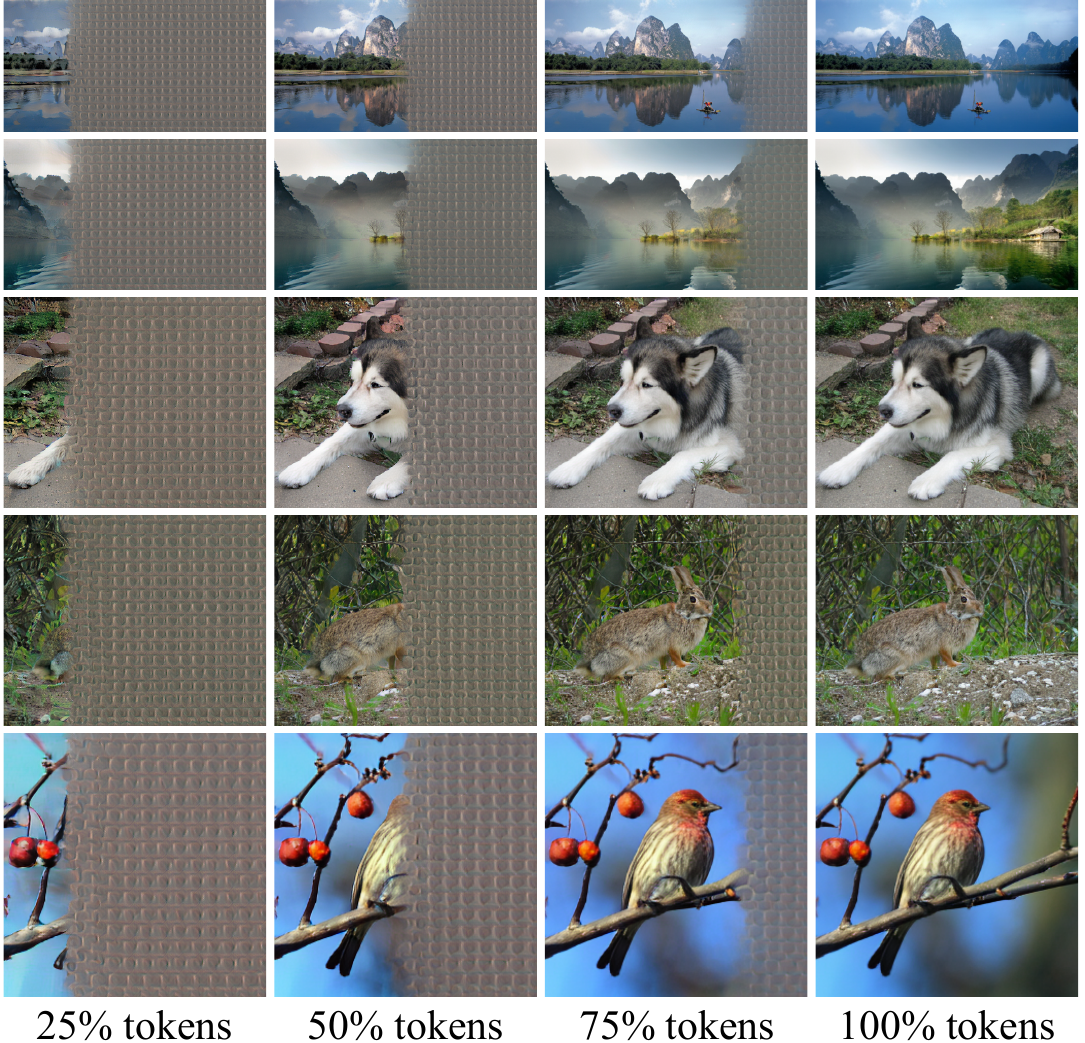}
    \vspace{-0.5cm}
    \caption{Additional Examples about Visual Meanings of 1D Tokens.}
    \label{fig:More_visual_tokens}
    \vspace{-0.5cm}
\end{figure}

\subsection{Tokenizer Training}
We follow the standard training recipe proposed in the Latent Diffusion Model (LDM)~\cite{rombach2022high}, and detailed hyper-parameter configurations used for training our equivariant 1D tokenizer are provided in ~\cref{tab:stage1}.

\subsection{Generative Models Training}

Detailed hyper-parameters utilized in our equivariant generator are summarized in ~\cref{tab:stage2}. To comprehensively evaluate model capacity and performance trade-offs, we trained multiple variants of the generative model with different sizes and complexities. The complete architectural details for each of these model variants are provided in ~\cref{tab:architecure}.

\begin{table*}[tp]
\centering
\begin{tabular}{l|cccccc}
\toprule
Model &  \#Para. & Layers & Hidden dim & Attn heads & Diff. hidden dim & Diff.layers\\
\midrule
Small & 151M & 16 & 512 & 8 & 960 & 12\\
Base & 294M & 24 & 768 & 12 & 1024 & 12\\
Large & 644M & 32 & 1024 & 16 & 1280 & 12\\
Huge & 1.2B & 40 & 1280 & 16 & 1536 & 12\\
\bottomrule
\end{tabular}
\caption{The model configurations of our generators. \#Para. denotes the number of parameters in the respective generators and Diff. presents the diffusion head. We also use "S", "B", and "L" and "H" as shorthand for different models in the manuscript.}
\label{tab:architecure}
\end{table*}

\begin{table}[tp]
\centering
\begin{tabular}{lc}
\toprule
config &  value \\
\midrule
Token length & 16 \\
Token channels & 256 \\
\midrule
MLP ratio & 4 \\
Norn layer in attention blocks & \textit{nn.LayerNorm }\\
Class labels sequence length & 16 \\
Class labels dropout & 0.1 \\
Attention dropout & 0.1 \\
Projection layer dropout & 0.1 \\
\midrule
Gradient clipping by norm & 3.0 \\
Optimizer & Adam \\
Beta1 & 0.9 \\
Beta2 & 0.95 \\
Base LR & 8.0e-4 \\
LR scheduler & constant \\
LR warmup epochs & 100 \\
Weight decay &  0.02\\
\midrule
EMA decay & 0.9999 \\
\midrule
Training epochs & 1200 \\
Total Batchsize & 2048 \\
GPU & A100 \\
\bottomrule
\end{tabular}
\caption{Detailed hyper-parameters for our equivariant generator.}
\label{tab:stage2}
\end{table}

We utilize 32 A100 GPUs for training the tokenizer, with the training process spanning 3 days. The generator models are trained using 64 A100 GPUs, requiring 4.6 days for the longest schedule (training our huge model for 1200 epochs).

\subsection{Sampling Hyper-parameters}
We generate results by sampling with 100 denoising steps, utilizing the first-order Euler solver following the Rectified Flow framework~\cite{liu2022flow}. At inference time, the genertor employs classifier-free guidance (CFG). Specifically, the underlying transformer network produces two distinct outputs: the conditional output $h_c$ (conditioning context present) and the unconditional output $h_u$ (conditioning context absent). The predicted velocity $v$ is obtained through interpolation of these two outputs as follows: $v=v_\theta(x_t,|t,h_u)+\omega \cdot (v_\theta(x_t,|t,h_c) - v_\theta(x_t,|t,h_u))$, where $\omega$ represents the guidance scale parameter. Inspired by Muse~\cite{chang2023muse}, we employ a dynamic CFG schedule in which the guidance scale $\omega$ increases linearly as the sampling sequence progresses. To maximize sampling quality, we systematically tune and select the optimal guidance scales individually for each trained model.

\section{Pseudo-Code for Our Equivariant 1D Tokenizer}\label{sec:pseudo}
We have included PyTorch-style pseudo-code for our Equivariant 1D Tokenizer.

\definecolor{commentcolor}{RGB}{110,154,155}   
\newcommand{\PyComment}[1]{\ttfamily\textcolor{commentcolor}{\# #1}}  
\newcommand{\PyCode}[1]{\ttfamily\textcolor{black}{#1}} 
\begin{algorithm*}[t]
\SetAlgoLined
    \PyCode{\textcolor{red}{class} EquivariantTokenizer(nn.Module)} \\
    \Indp \PyCode{\textcolor{red}{def} \_\_init\_\_(token\_channels):} \\
    \Indp \PyComment{2D Encoder and Decoder} \\
    \PyCode{self.2DEncoder, self.2DDecoder = 2DEncoder(), 2DDecoder()} \\
    \PyComment{1D Encoder and Decoder} \\
    \PyCode{self.1DEncoder, self.1DDecoder = 1DEncoder(), 1DDecoder()} \\
    \PyCode{self.token\_channels = token\_channels} \\
    \PyCode{} \\
    
    \Indm \PyCode{\textcolor{red}{def} forward(self, x):} \\
    \Indp\PyCode{z = self.2DEncoder(x)}\\

    \PyCode{} \\
    \PyComment{Columnization}\\
    \PyCode{z = z.permute(0,3,1,2)} \\
    \PyCode{z = z.reshape(z.shape[0], z.shape[1], -1)}\\

    \PyCode{} \\
    \PyComment{1D Latent}\\
    \PyCode{posterior = self.1DEncoder(x)} \\
    \PyCode{latent = posterior.sample()} \\
    \PyCode{z = self.1DDecoder(latent)}\\

    \PyCode{} \\
    \PyComment{Rasterization}\\
    \PyCode{z = z.reshape(z.shape[0], z.shape[1], self.token\_channels, -1)}\\
    \PyCode{z = z.permute(0, 2, 3, 1)}\\

    \PyCode{} \\
    \PyCode{\textcolor{red}{return} self.2DDecoder(z)}
    
\caption{Our Equivariant 1D Tokenizer PyTorch-style Pseudo-Code}
\label{algo:tokenizer}
\end{algorithm*}

\begin{figure}[tp]
    \centering
    \includegraphics[width=\linewidth]{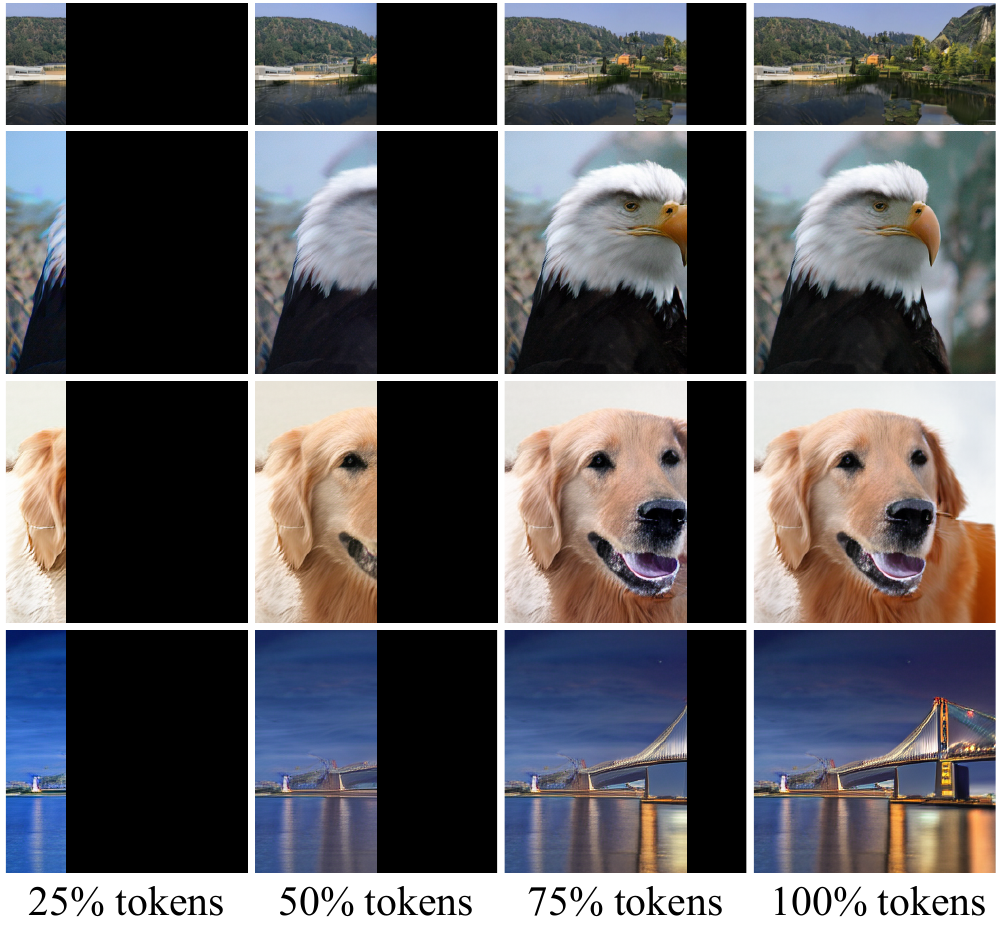}
    \vspace{-0.5cm}
    \caption{Visualization of the generation process. }
    \label{fig:generation_process}
\end{figure}

\section{Visualization of Our Huge Model on ImageNet}\label{sec:imagenet}
\begin{figure*}
    \centering
    \includegraphics[width=\linewidth]{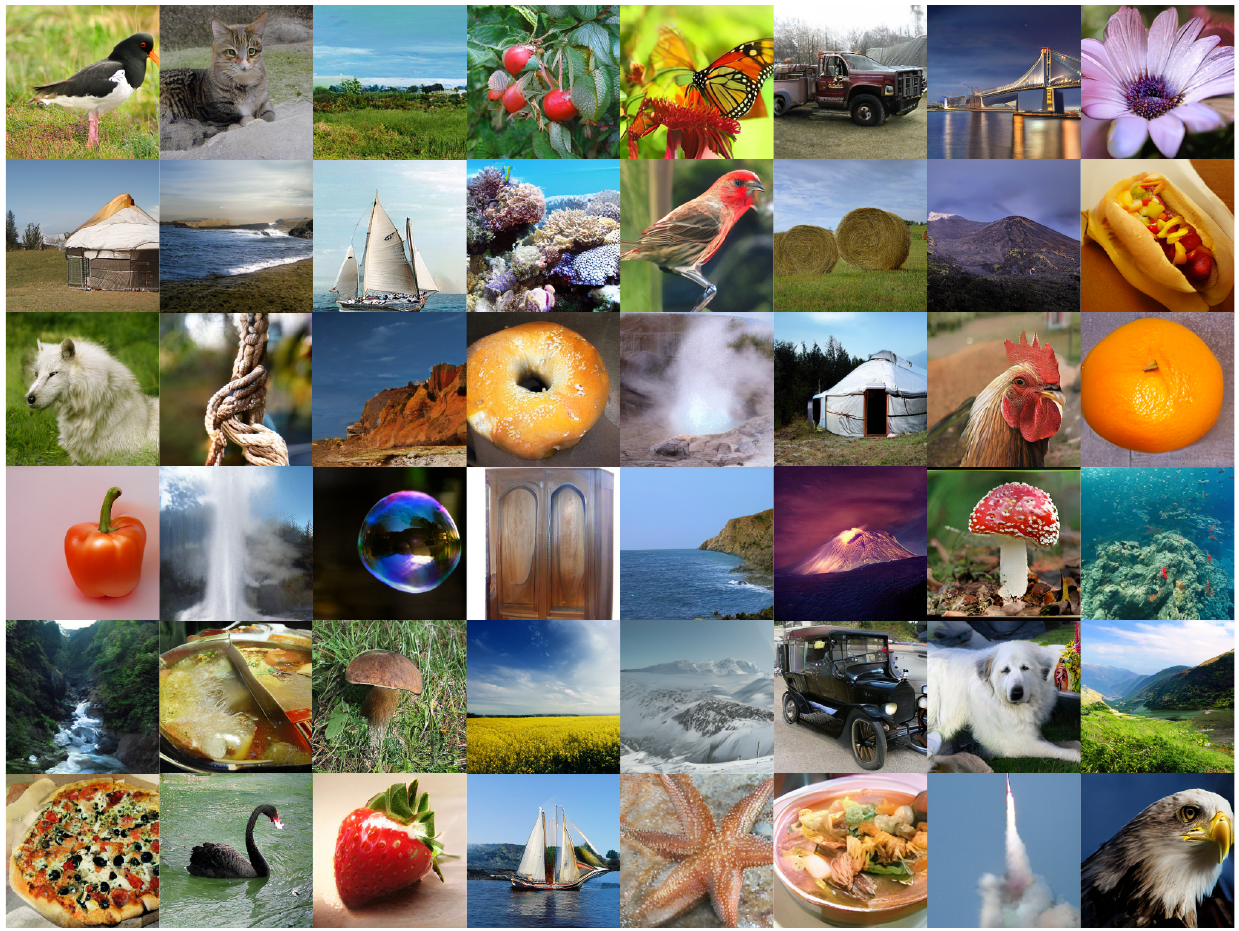}
    \caption{Generation Results on the ImageNet-1k Dataset.}
    \label{fig:imagenet}
\end{figure*}

We showcase the uncurated 256×256 images generated by our huge model in ~\cref{fig:imagenet}.

\section{Visualization of Long Images}\label{sec:places}
\begin{figure*}
    \centering
    \includegraphics[width=0.9\linewidth]{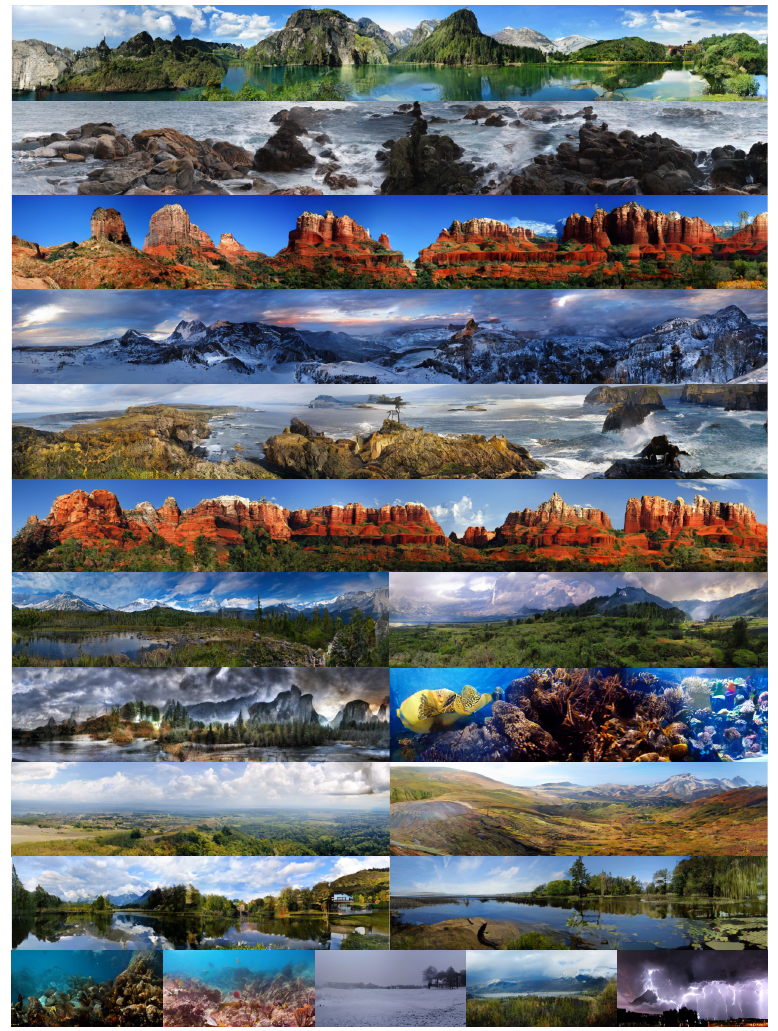}
    \caption{More Visual Examples of Generated Long Images.}
    \label{fig:More Long Images}
\end{figure*}

We provide uncurated long-content images produced by our large model in ~\cref{fig:More Long Images}, which is trained exclusively on the Nature subset of the Places dataset. Owing to the equivariant property, our model effectively captures fine-grained spatial coherence, enabling it to generate high-fidelity landscape images. Remarkably, our approach demonstrates this zero-shot capability, as the model was never explicitly trained on high-resolution images.

\section{Visualization of Tokens}\label{sec:token_meaning}
To clearly illustrate the relationship between encoded tokens and corresponding visual content, we progressively replace the randomly initialized token sequences with encoded tokens. As demonstrated in ~\cref{fig:Spatial decoupling} and further substantiated in~\cref{fig:More_visual_tokens}, the decoder faithfully reconstructs the original images step by step. Furthermore, we visualize the generation process by decoding progressively generated token sequences in ~\cref{fig:generation_process}, thereby providing further clarification regarding the semantic interpretation of tokens in the latent space from a generative perspective.
\end{document}